\documentclass[11pt]{article}

\usepackage{acl}

\usepackage{times}
\usepackage{latexsym}

\usepackage[T1]{fontenc}

\usepackage[utf8]{inputenc}

\usepackage{microtype}

\usepackage{inconsolata}

\usepackage{graphicx}
\usepackage{tabularx}
\usepackage{booktabs}
\usepackage{makecell}
\usepackage{pifont}
\usepackage{amsmath, amssymb}
\usepackage{fontawesome5}
\usepackage{algorithm, algpseudocode}
\usepackage{pgfplots}
\pgfplotsset{compat=1.18}
\usepackage{subcaption}
\usepackage{listings}
\usepackage{here}

\usepackage[svgnames]{xcolor}       
\usepackage[many]{tcolorbox}        

\lstdefinestyle{prompt}{
    literate={`}{\textasciigrave}{1},
    basicstyle=\ttfamily\small\selectfont\color{gray!32!black},
    backgroundcolor=\color{cyan!5},
    rulecolor=\color{cyan!30},
    frame=single,
    framerule=0.8pt,
    framexleftmargin=0pt,
    framexrightmargin=0pt,
    framextopmargin=0pt,
    framexbottommargin=0pt,
    breaklines=true,
    breakindent=0pt,
    keepspaces=true,
    tabsize=2, 
    xleftmargin=0pt,
    xrightmargin=0pt,
    aboveskip=0pt,
    belowskip=0pt,
    commentstyle=\color{gray!65}\itshape,
    keywordstyle=\bfseries\color{teal!70}
}

\title{Agentar-Scale-SQL: Advancing Text-to-SQL through Orchestrated Test-Time Scaling}

\author{Pengfei Wang$^{1*}$, Baolin Sun$^{1* \dag}$, Xuemei Dong$^{1*}$, Yaxun Dai$^{2* \ddagger}$, Hongwei Yuan$^{3* \ddagger}$, \\ {\bf Mengdie Chu$^{1}$, Yingqi Gao$^{1}$, Xiang Qi$^{1 \dag}$, Peng Zhang$^{1 \dag}$, Ying Yan$^{1}$} \\
        $^{1}$Ant Digital Technologies, Ant Group, $^{2}$Soochow University, $^{3}$Zhejiang University \\ \{nanzhou.wpf, xuanfeng.sbl, dongxuemei.dxm, qixiang.qx, minghua.zp\}@antgroup.com \\ \faGithub ~\url{https://github.com/antgroup/Agentar-Scale-SQL} \\}

\begin{document}
\maketitle

\begin{abstract}
\def\thefootnote{*}\footnotetext{Core contributors.}
\def\thefootnote{$\ddagger$}\footnotetext{This work was partially done during the author's internship at Ant Digital Technologies, Ant Group.}
\def\thefootnote{$\dag$}\footnotetext{Corresponding authors.}

State-of-the-art (SOTA) Text-to-SQL methods still lag significantly behind human experts on challenging benchmarks like BIRD. Current approaches that explore test-time scaling lack an orchestrated strategy and neglect the model's internal reasoning process. To bridge this gap, we introduce Agentar-Scale-SQL, a novel framework leveraging scalable computation to improve performance. Agentar-Scale-SQL implements an Orchestrated Test-Time Scaling strategy that synergistically combines three distinct perspectives: i) Internal Scaling via RL-enhanced Intrinsic Reasoning, ii) Sequential Scaling through Iterative Refinement, and iii) Parallel Scaling using Diverse Synthesis and Tournament Selection. Agentar-Scale-SQL is a general-purpose framework designed for easy adaptation to new databases and more powerful language models. Extensive experiments show that Agentar-Scale-SQL achieves SOTA performance on the BIRD benchmark, reaching 81.67\% execution accuracy on the test set and ranking first on the official leaderboard, demonstrating an effective path toward human-level performance.

\end{abstract}

\section{Introduction}

Democratizing access to data analytics by enabling users to query structured databases in their own natural language is a long-standing goal in human-computer interaction. This is the core objective of Text-to-SQL, a pivotal research area focused on translating natural language questions into executable SQL queries \citep{DBLP:journals/corr/abs-2403-02951, DBLP:journals/pvldb/LiLCLT24, DBLP:journals/corr/abs-2408-05109, DBLP:journals/pvldb/LuoLFCT25}. By bridging the gap between human language and structured data, Text-to-SQL empowers non-technical users to interact with complex databases effectively, garnering significant interest from both the Natural Language Processing (NLP) and database communities \citep{DBLP:conf/nips/LiHQYLLWQGHZ0LC23, DBLP:conf/emnlp/YuZYYWLMLYRZR18, DBLP:conf/iclr/PourrezaL0CTKGS25, DBLP:journals/corr/abs-2505-04671}.

The ultimate vision for Text-to-SQL is to develop systems that can match, and eventually surpass, human expert performance. However, a significant gap persists between this ambition and the current state-of-the-art. On the challenging BIRD benchmark \citep{DBLP:conf/nips/LiHQYLLWQGHZ0LC23}, human experts achieve an execution accuracy (EX) of 92.96\%, whereas even the top-performing methods lag considerably, with the top 5 approaches around 75\% on the test set.
Closing the vast human-machine performance divide urgently requires innovation.

Recent Text-to-SQL research falls into three main categories. The first consists of prompt-based methods (e.g., OpenSearch-SQL \citep{DBLP:journals/pacmmod/XieXZG25} and DAIL-SQL \citep{DBLP:journals/pvldb/GaoWLSQDZ24}). The second category is comprised of fine-tuning-based approaches, with representative works like Arctic-Text2SQL-R1-32B \citep{DBLP:journals/corr/abs-2505-20315}. 
The third category consists of hybrid methods like XiYan-SQL \citep{DBLP:journals/corr/abs-2507-04701}, CHASE-SQL + Gemini \citep{DBLP:conf/iclr/PourrezaL0CTKGS25}, and Contextual-SQL \citep{agrawal2025text2sql}. These approaches primarily explore test-time scaling from different perspectives: Contextual-SQL adopts an ensemble strategy, while XiYan-SQL and CHASE-SQL investigate ensemble strategies and sequential refinement. However, these studies share a common limitation, as they neglect both an internal perspective on the model's reasoning process and the orchestrated scaling combination.

\begin{figure*}[ht]
    \centering
    \includegraphics[width=6.3in]{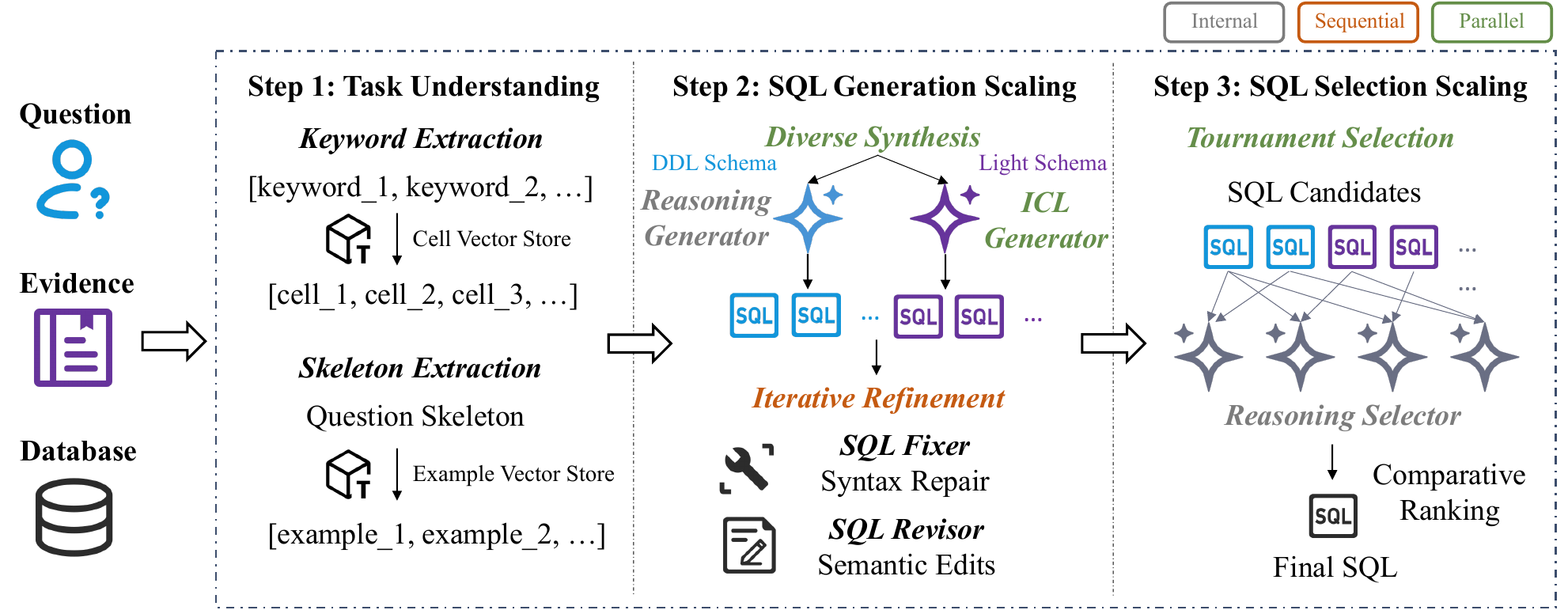}
    \caption{The proposed Agentar-Scale-SQL framework.}
    \label{fig:framework}
    \vspace{-2mm}
\end{figure*}
To further advance Text-to-SQL performance, this work contends that the most promising path forward lies in fully embracing the principle of "The Bitter Lesson" \citep{sutton2019bitter}: that general methods leveraging scalable computation ultimately triumph over those based on complex, human-specified knowledge. With this philosophy, we focus on test-time scaling and have not designed complex strategies for schema linking, as we believe that scaling alone is sufficient to improve performance.
To this end, we introduce Agentar-Scale-SQL, an \emph{Orchestrated Test-Time Scaling} framework that demonstrates that the path to human-level performance lies not in crafting more intelligent heuristics but in building general-purpose frameworks that effectively leverage the test-time scaling \citep{DBLP:journals/corr/abs-2408-03314, DBLP:journals/corr/abs-2503-24235}.
Agentar-Scale-SQL employs an orchestrated scaling strategy across three distinct perspectives: i) Internal Scaling (through RL-enhanced Intrinsic Reasoning), ii) Sequential Scaling (through Iterative Refinement), and iii) Parallel Scaling (through both Diverse Synthesis and Tournament Selection).
This strategy is realized within a three-stage framework, as depicted in Figure \ref{fig:framework}.
The objective of Step 1 (Task Understanding) is to establish a comprehensive understanding of the input and its context, which is essential for the scaling operations in subsequent steps.
Then, Step 2 (SQL Generation Scaling) develops diverse synthesis and iterative refinement to obtain high-quality and diverse SQL candidates.
Specifically, diverse synthesis employs two distinct generators (i.e., intrinsic reasoning generator and ICL generator) with generator-specific schema formats.
Finally, Step 3 (SQL Selection Scaling) employs tournament selection and fine-tunes an intrinsic reasoning selector to ensure high accuracy.
We summarize our contributions as follows:
\begin{itemize}
    \item \emph{Orchestrated Test-time Scaling.} We introduce an orchestrated test-time scaling framework that converts additional inference compute into accuracy gains by scaling across three distinct perspectives: i) Internal Scaling (RL-enhanced Intrinsic Reasoning), ii) Sequential Scaling (Iterative Refinement), and iii) Parallel Scaling (Diverse Synthesis and Tournament Selection). 
    \item \emph{General Purpose and Scalability.} The entire framework is fully general-purpose, plug-and-play, and effortlessly adaptable to any database. As future LLMs become more powerful and computing continues to get cheaper, Agentar-Scale-SQL’s performance ceiling will lift on its own; we can simply allocate more compute to both generation and selection stages to achieve even higher accuracy.
    \item \emph{Transparent and Actionable Insights.} We propose a structured three-stage Text-to-SQL framework, delineating the specific roles and objectives of each stage. Crucially, our work is the first to scale both the SQL generation and selection stages. Together, these contributions establish transparent and actionable insights to guide future research and development in the Text-to-SQL field.
    \item \emph{Extensive Experiments.} Extensive experiments confirm the SOTA performance of Agentar-Scale-SQL. Specifically, Agentar-Scale-SQL achieves an EX of 74.90\% on the BIRD development set and 81.67\% on the test set, along with an R-VES of 77.00\%. With these results\footnote{See the official BIRD leaderboard at \url{https://bird-bench.github.io/}. The rankings are dated November 27, 2025.}, Agentar-Scale-SQL \textbf{ranks first} on the BIRD leaderboard. Our findings indicate that scaling is a critical factor for achieving SOTA performance in Text-to-SQL.
\end{itemize}

\section{Related Work}
\noindent \textbf{Text-to-SQL.}
While LLMs have significantly advanced Text-to-SQL, their direct application remains challenging for complex queries \citep{DBLP:journals/corr/abs-2408-05109, DBLP:journals/tkde/LiuSLMJZFLTL25}. Recent approaches address this via prompt-based methods \citep{DBLP:journals/pacmmod/XieXZG25, DBLP:journals/pvldb/GaoWLSQDZ24, DBLP:journals/corr/abs-2307-07306}, fine-tuning \citep{DBLP:journals/corr/abs-2505-20315, DBLP:journals/corr/abs-2503-23157, DBLP:journals/corr/abs-2503-02240, DBLP:journals/pacmmod/LiZLFZZWP0024,DBLP:conf/acl/YangHY0LZ24}, or hybrid strategies that employ test-time scaling \citep{DBLP:journals/corr/abs-2507-04701, DBLP:conf/iclr/PourrezaL0CTKGS25, agrawal2025text2sql}. Most relevant to our work are these scaling methods, which use techniques such as parallel synthesis and sequential refinement to improve generation quality. However, these approaches share a common limitation in that they lack an orchestrated combination of different scaling dimensions.

\noindent \textbf{Test-time Scaling.}
Test-time Scaling (TTS) enhances an LLM's performance during inference by strategically expanding computation, rather than altering model weights \citep{DBLP:journals/corr/abs-2503-24235, DBLP:journals/corr/abs-2408-03314, DBLP:journals/corr/abs-2001-08361}. Existing research in TTS can be broadly classified into three paradigms. Parallel scaling focuses on generating multiple candidate solutions concurrently and aggregating them to improve the likelihood of finding a correct answer, with self-consistency \citep{DBLP:conf/iclr/0002WSLCNCZ23} being a prominent example. In contrast, sequential scaling emulates a deliberative "System 2" reasoning process by iteratively building or refining a solution through a series of steps, as exemplified by Chain-of-Thought (CoT) \citep{DBLP:conf/nips/Wei0SBIXCLZ22} and iterative refinement \citep{DBLP:conf/nips/MadaanTGHGW0DPY23}. More recently, internal scaling has emerged, where the model is trained to autonomously allocate its own computational budget and determine reasoning depth without external orchestration \citep{DBLP:journals/corr/abs-2501-12948, DBLP:journals/corr/abs-2412-16720}. Our work, Agentar-Scale-SQL, builds upon these foundational concepts by introducing a novel orchestration framework that synergistically combines the three paradigms, specifically tailored to advance SOTA in the Text-to-SQL domain.

\section{Methodology}

\subsection{Overview}

Given a question $Q_u$, evidence $E_u$, and a target database $D$, our framework is divided into three stages, as depicted in Figure \ref{fig:framework}.
An offline preprocessing phase precedes the operation of the online framework.

\noindent \textbf{Offline Preprocessing.}
Our method involves three offline preprocessing steps before online inference. First, to increase the diversity of generation, we represent the database metadata in two formats (see Figure \ref{fig:light_schema}). 
We create a Markdown-based light schema designed for effective In-Context Learning (ICL) with general-purpose LLMs and use a standard DDL schema to fine-tune the code-specialized model, which capitalizes on its training background to achieve faster convergence. Second, we index all textual cell values from the database into a vector store ${VD}_{cell}$. Finally, we also index the training set as examples into a vector store ${VD}_{example}$, which allows us to retrieve relevant few-shot examples during inference by embedding the skeleton-extracted user question and performing a similarity search.

\noindent \textbf{Online Framework.}
Agentar-Scale-SQL is an orchestrated test-time scaling framework that converts additional inference compute into accuracy gains by employing an orchestrated scaling strategy across three distinct perspectives: i) Internal Scaling (through RL-enhanced Intrinsic Reasoning), ii) Sequential Scaling (through Iterative Refinement), and iii) Parallel Scaling (through both Diverse Synthesis and Tournament Selection).
Overall, Agentar-Scale-SQL consists of three stages: Step 1 (Task Understanding), Step 2 (SQL Generation Scaling), and Step 3 (SQL Selection Scaling).
\begin{itemize}
    \item \textbf{Step 1 (Task Understanding)} focuses on comprehensively understanding the user's intent and retrieving relevant context. 
    \item \textbf{Step 2 (SQL Generation Scaling)} develops diverse synthesis and iterative refinement to obtain high-quality and diverse SQL candidates.
    The diverse synthesis component, in particular, utilizes two distinct generators operating on a specific schema format: a reasoning generator with DDL schema and an ICL generator with light schema.
    \item \textbf{Step 3 (SQL Selection Scaling)} utilizes a tournament selection method, enhanced by an intrinsic reasoning selector, to achieve high selection accuracy.
\end{itemize}

The following sections delve into the details of each component.

\subsection{Task Understanding}

Database cells are crucial \citep{DBLP:conf/iclr/PourrezaL0CTKGS25, DBLP:journals/corr/abs-2507-04701}, as they provide the specific values needed for SQL clauses like WHERE and HAVING. Similarly, well-chosen few-shot examples are known to significantly improve the performance of In-Context Learning (ICL) \citep{DBLP:journals/pvldb/GaoWLSQDZ24}. Therefore, the primary objective of the Task Understanding step is to identify and retrieve these two critical forms of context: relevant database cells and effective demonstration examples.
This is achieved through two parallel sub-processes: i) keyword extraction that extracts keywords from the question $Q_u$ and evidence $E_u$ to retrieve relevant cells using embedding-based similarity from ${VD}_{cell}$, ii) skeleton extraction from the $Q_u$ to retrieve relevant examples using embedding-based similarity from ${VD}_{example}$.

\subsection{SQL Generation Scaling}

\begin{figure}
    \centering
    \includegraphics[width=3in]{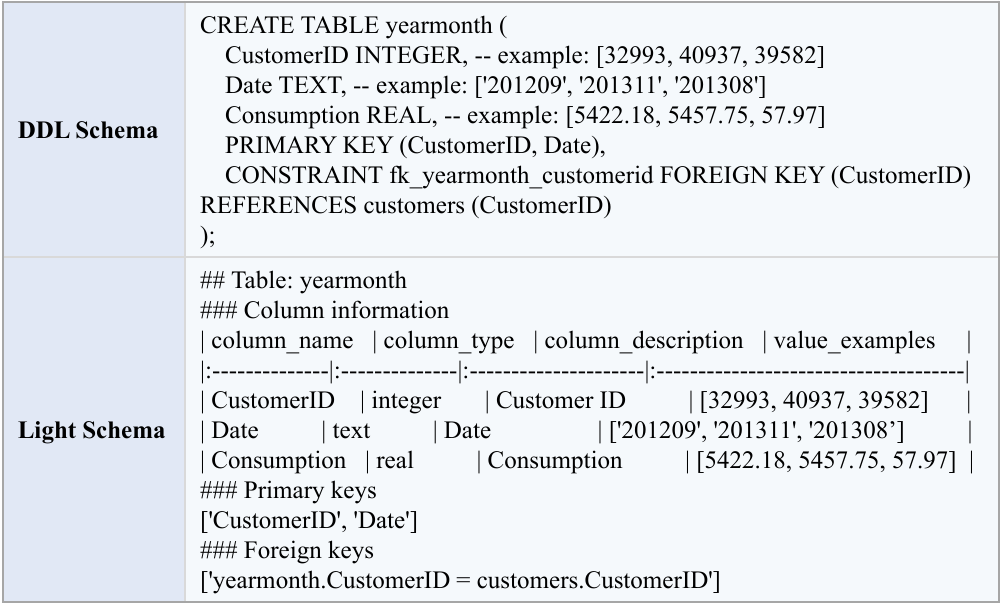}
    \caption{An example of a database schema represented in both DDL schema and light schema formats.}
    \label{fig:light_schema}
\end{figure}

This stage employs two complementary generators, operating on dual schema views, to produce high-quality and diverse candidates. The first generator, ${M}_{reasoning}$, is an intrinsic reasoner trained via reinforcement learning (RL). The second, ${M}_{ICL}$, is an in-context learning (ICL) model driven by a large, proprietary LLM.
Then, an iterative refinement loop for syntax repair and semantic edits.
As a result, a set of $n$ SQL candidates, denoted as $C=\{c_{1}, {c_2}, ..., c_{n} \}$, is generated.
The following sections detail each component.

\subsubsection{Intrinsic Reasoning SQL Generator}
\label{sec:reasoning_generator}
Inspired by Arctic-Text2SQL-R1 \citep{DBLP:journals/corr/abs-2505-20315}, we pursue robust intrinsic reasoning Text-to-SQL generation via a simple, execution-grounded RL framework.

\noindent \textbf{Overview of RL Pipeline.}
We adopt GRPO \citep{DBLP:journals/corr/abs-2402-03300} due to its proven efficiency and effectiveness on structured reasoning tasks. 
Unlike Arctic-Text2SQL-R1, we exclusively use the training data from the BIRD dataset, without any extra data.
Formally, let $\pi_\theta$ denote our policy model parameterized by $\theta$. For any given input question $Q_u$, its associated evidence and schema, the model generates a set of $N$ candidate SQL queries (i.e., rollouts), $\{o_{Q,1}, \dots, o_{Q,N}\}$. Each candidate is then evaluated to yield an explicit reward signal, as described later. This per-input batch of rollouts allows for computing relative advantages, thereby stabilizing learning and fostering robust policy updates.

The clipped surrogate objective for each sample $i$ is defined as:
\begin{equation}
\begin{aligned}
    L(\theta) ={}& \frac{1}{N}\sum_{i=1}^{N} \min\left(r_i A_i, \right. \\
                 & \qquad \left. \text{clip}(r_i, 1-\epsilon, 1+\epsilon)A_i\right)
\end{aligned}
\end{equation}
The full GRPO objective:
\begin{equation}
    \mathcal{J}_{\text{GRPO}}(\theta) = \mathbb{E}[L(\theta)] - \beta D_{\text{KL}}(\pi_{\theta} \| \pi_{\text{ref}})
\end{equation}

where $r_i$ is the likelihood ratio $\frac{\pi_\theta(o_i|Q)}{\pi_{\theta_{\text{old}}}(o_i|Q)}$, $A_i$ is the advantage, and $D_{\text{KL}}$ is a KL-divergence penalty that keeps the policy close to a reference (supervised fine-tuned) model \citep{DBLP:conf/nips/Ouyang0JAWMZASR22}. The parameters $\epsilon$ and $\beta$ are tuned in practice to balance exploration and stability.

\noindent \textbf{Reward Design.}
We define a reward function $R_G$ solely on final execution correctness and basic syntax validity following Arctic-Text2SQL-R1:
\begin{equation}
R_G = 
\begin{cases} 
    1,   & \text{if results match the ground truth;} \\
    0.1, & \text{if the SQL is executable;} \\
    0,   & \text{otherwise.}
\end{cases}
\end{equation}

\subsubsection{Diverse Synthesis}
The diverse synthesis strategy is designed to generate a high-quality and varied pool of SQL candidates. This strategy involves two parallel and complementary generators: a fine-tuned reasoning generator and an ICL generator.
We create a Markdown-based light schema designed for effective ICL with general-purpose LLMs and use a standard Definition Language (DDL) schema to fine-tune the code-specialized model, which capitalizes on its training background to achieve faster convergence.

\noindent \textbf{Reasoning Generator.}
This generator utilizes the DDL schema to conduct deep, step-by-step reasoning. On the one hand, guided by our Internal Scaling principle, it is engineered to construct complex queries with the primary goal of achieving high accuracy. On the other hand, it can be fine-tuned to align with the specific characteristics and requirements of the target benchmark.

\noindent \textbf{ICL Generator.}
In parallel, the ICL Generator utilizes a Markdown-based light schema (see Figure \ref{fig:light_schema}) and the few-shot examples retrieved during the Task Understanding stage. 
To enhance the diversity of the ICL generator, we employ a multi-faceted strategy: varying the input prompts, randomizing the order of in-context examples, utilizing a range of LLMs, and adjusting the temperature settings.
Typically, the varied prompts are categorized into three distinct styles: direct prompting (without explicit reasoning), Chain-of-Thought (CoT) prompting \citep{DBLP:conf/nips/Wei0SBIXCLZ22}, and problem decomposition.
This generator excels at rapidly producing a broad range of plausible queries by leveraging pattern recognition from the provided in-context examples.

By combining the deep, analytical approach of the reasoning generator with the example-driven approach of the ICL generator, we maximize the diversity of the candidate pool. This synergy significantly increases the probability that at least one correct or near-correct query is present before the selection phase.

\subsubsection{Iterative Refinement}
To further improve the quality of SQL candidates, we introduce the iterative refinement module to refine errors.
SQL queries can contain both syntactic and semantic errors \citep{DBLP:journals/pacmmod/YangWXWDPCL25, DBLP:conf/coling/XuLJDSWX25}. 
We employ a two-pronged approach to refine errors. For syntactic errors (the former), we use the SQL fixer, an LLM-based component that is conditionally activated to repair invalid syntax. For semantic errors (the latter), we employ the SQL revisor, an LLM agent designed to identify and refine logical flaws in the query.
To streamline the revision process, we first group queries by their execution outcomes. Subsequently, we randomly select one query from each group for refinement.

\subsection{SQL Selection Scaling}

The primary limitation of majority voting is its underlying assumption that the most frequent answer is also the correct one, a premise that does not always hold.
Instead, we employ a tournament selection process where a reasoning selector, enhanced by reinforcement learning (RL), evaluates candidates through pairwise comparisons. The top-ranked SQL query from this process is selected as the final SQL.
We detail these modules in the following sections.

\subsubsection{Tournament Selection}
We select an optimal SQL query in a two-stage process. First, we consolidate an initial pool of queries by grouping them based on identical execution results on a database $D$. A single representative is chosen from each group to form a candidate set $C' = \{c_1, \dots, c_m\}$. Second, these candidates compete in a pairwise round-robin tournament. For each pair $(c_i, c_j)$, a reasoning selector $\mathcal{M}_{selection}$ determines a winner based on the question, light schema, and execution results, incrementing the winner's score $W_i$. The final query is the one with the maximum score: $c_{final} = \underset{c_i \in C'}{\arg\max} \, W_i$.

\subsubsection{Intrinsic Reasoning SQL Selector}
We apply Reinforcement Learning (RL) to SQL selector $\mathcal{M}_{selection}$, an approach analogous to the intrinsic reasoning used in the SQL generator.

\noindent \textbf{Overview of RL Pipeline.}
Following the methodology described in Section \ref{sec:reasoning_generator}, we apply GRPO to enhance the reasoning capabilities for SQL selection.
Based on the training set, we construct 8.5k samples for reinforcement learning.

\noindent \textbf{Reward Design.}
The objective of SQL selection is to identify the correct query from a set of candidates. To achieve this, we introduce a result-oriented reward function, $R_S$, designed to evaluate the correctness of the selection:
\begin{equation}
R_S = 
\begin{cases} 
    1,   & \text{if the selection is correct;} \\
    0,   & \text{otherwise.}
\end{cases}
\end{equation}

\section{Experiments}
In this section, we experimentally evaluate the proposed Agentar-Scale-SQL on the large-scale dataset. We aim to answer the following research questions:
\begin{itemize}
    \item \textbf{RQ1:} How does Agentar-Scale-SQL perform compared with the state-of-the-art methods?
    \item \textbf{RQ2:} How does each module affect the overall performance of Agentar-Scale-SQL?
    \item \textbf{RQ3:} What are the individual and complementary roles of the ICL and Reasoning generators?
    \item \textbf{RQ4:} How does performance scale with the number of candidates across different complexity levels?
\end{itemize}

\subsection{Experimental Setup}

\noindent \textbf{Datasets.}
We evaluate our method on the BIRD benchmark \citep{DBLP:conf/nips/LiHQYLLWQGHZ0LC23}, a particularly challenging cross-domain dataset. It comprises over 12,751 question-SQL pairs across 95 large databases, simulating real-world complexity with messy data and intricate schemas across more than 37 professional domains.

\noindent \textbf{Baselines.}
We compared several top-ranking baseline methods from the overall leaderboard and the single-model leaderboard.
The former consists of fifteen baselines, including AskData + GPT-4o \citep{DBLP:journals/corr/abs-2505-19988}, LongData-SQL, CHASE-SQL + Gemini \citep{DBLP:conf/iclr/PourrezaL0CTKGS25}, JoyDataAgent-SQL, XiYan-SQL \citep{DBLP:journals/corr/abs-2507-04701}, among others.
The latter comprises eight leading methods, such as Gemini-SQL, Databricks RLVR 32B, Sophon-Text2SQL-32B, Arctic-Text2SQL-R1-32B \citep{DBLP:journals/corr/abs-2505-20315}.

\noindent \textbf{Metrics.}
Following prior work \citep{DBLP:conf/iclr/PourrezaL0CTKGS25}, we use Execution Accuracy (EX), the official metric for the respective leaderboard, as the primary evaluation metric to compare methods.
Besides, we adopt the official Reward-based Valid Efficiency Score (R-VES) to evaluate the efficiency of the generated SQL.

\noindent \textbf{Implementation Details.}
We implement Agentar-Scale-SQL with LangChain\footnote{\url{https://github.com/langchain-ai/langchain}}/LangGraph\footnote{\url{https://github.com/langchain-ai/langgraph}} and chroma\footnote{\url{https://github.com/chroma-core/chroma}} retrieval using all-MiniLM-L6-v2\footnote{\url{https://huggingface.co/sentence-transformers/all-MiniLM-L6-v2}} embeddings.
The Task Understanding is powered by Gemini-2.5-Flash \citep{DBLP:journals/corr/abs-2507-06261} (temperature 0.2). The ICL SQL Generator utilizes Gemini-2.5-Pro \citep{DBLP:journals/corr/abs-2507-06261} (abbreviated as pro) with two temperature settings (0.5 and 1.8) and GPT-5 \citep{gpt-5} (minimal reasoning effort).
The Reasoning SQL Generator is fine-tuned based on Omni-SQL-32B \citep{DBLP:journals/corr/abs-2503-02240}. By default, candidates comprise 9 from the ICL SQL Generator and 8 from the Reasoning SQL Generator. 
The SQL Fixer and SQL Reviser both use pro.
The base model of the Reasoning SQL Selector is Qwen2.5-Coder-32B-Instruct \citep{DBLP:journals/corr/abs-2409-12186}.
The RL framework leverages verl \citep{DBLP:conf/eurosys/ShengZYWZZPL025} on 32 NVIDIA A100 80GB GPUs.

\begin{table*}[ht]
\centering
\caption{Evaluation results on the development and test sets.}
\label{table:main_bird}
\begin{tabular}{cccc}
\toprule
\textbf{Methods} & \textbf{EX (Dev)} & \textbf{EX (Test)} & \textbf{R-VES (\%)} \\ \midrule
\multicolumn{4}{c}{\textbf{Overall}}                                                  \\
\midrule
Alpha-SQL \citep{DBLP:journals/corr/abs-2502-17248}    &  69.70         &  70.26                  &      -               \\
OmniSQL-32B \citep{DBLP:journals/corr/abs-2503-02240}      &   69.23      &    72.05          &  67.05               \\
OpenSearch-SQL \citep{DBLP:journals/pacmmod/XieXZG25}    &    69.30        &  72.28     & 69.36                    \\
Reasoning-SQL 14B \citep{DBLP:journals/corr/abs-2503-23157}              &      72.29             &     72.78               &   68.67                  \\
ExSL + granite-34B-code        &    72.43               &    73.17                &   71.37                  \\
CSC-SQL \citep{DBLP:journals/corr/abs-2505-13271}     &     71.33       &      73.67              &  67.84                   \\
CYAN-SQL         &   73.47           &    75.35                &     -                \\
XiYan-SQL \citep{DBLP:journals/corr/abs-2507-04701}        &     73.34              &   75.63                 &    71.41                 \\
Contextual-SQL \citep{agrawal2025text2sql}   &     73.50          &     75.63               &   70.02                  \\
TCDataAgent-SQL     &       74.12        &    75.74       &      -               \\
JoyDataAgent-SQL     &       74.25        &    75.85       &      70.16               \\
CHASE-SQL + Gemini \citep{DBLP:conf/iclr/PourrezaL0CTKGS25}                 &     74.90        &   76.02            &   69.94                  \\
LongData-SQL         &      74.32         &        77.53            &     71.89                \\ 
AskData + GPT-4o \citep{DBLP:journals/corr/abs-2505-19988}         &   76.14           &    80.88                &   76.24                  \\
\midrule
\multicolumn{4}{c}{\textbf{Single Trained Model}}                                                  \\
\midrule
SHARE \citep{DBLP:conf/acl/QuLQLH0C25}        &          64.14     &      -              &        -             \\
Syn CoT + DPO \citep{DBLP:conf/acl/LiuLZCXTQZ25}       &          67.10     &      -              &        -             \\
XiYanSQL-QwenCoder-32B \citep{DBLP:journals/corr/abs-2507-04701}        &          67.01     &      69.03              &        -             \\
Jiayin-Pangu-Text2SQL-14B     &    71.10       &   73.45          &    -                 \\
Arctic-Text2SQL-R1-32B \citep{DBLP:journals/corr/abs-2505-20315}    &   72.20            &     73.84               &      -               \\
Sophon-Text2SQL-32B      &     72.43          &     74.79         &      -               \\
Databricks RLVR 32B \citep{ali2025stateoftheartsqlreasoningmodel}      &      -      &    75.68              &     -                \\
Gemini-SQL (Multitask SFT + Gemini-2.5-Pro)      &      72.62      &    76.13              &     -                \\\midrule
\multicolumn{4}{c}{\textbf{Ours}}                                                  \\
\midrule
Agentar-Scale-SQL (Ours)        &   \textbf{74.90}              &     \textbf{81.67}             &    \textbf{77.00}               \\ \bottomrule
\end{tabular}

\end{table*}

\begin{table*}[htbp]
\centering
\caption{Ablation results on the development set.}
\label{table:ablation}
\begin{tabular}{cccccc}
\toprule
\textbf{Methods} & \textbf{Simple} & \textbf{Moderate} & \textbf{Challenging} & \textbf{Total} & \textbf{$\Delta$Total} \\ \midrule
    Agentar-Scale-SQL             &   79.35           &    69.40      &   64.14      &    74.90 &  -  \\ \midrule
    \textbf{w/o} Task Understanding & 79.14 & 68.32 & 64.14 & 74.45 & -0.45 \\
          \textbf{w/o} Reasoning SQL Generator & 74.92 & 63.79 & 58.62 & 70.01 & -4.89 \\
          \textbf{w/o} ICL SQL Generator & 75.89 & 66.38 & 55.86 & 71.12 & -3.78 \\ 
          \textbf{w/o} Iterative Refinement & 78.92 & 68.75 & 63.45 & 74.38 & -0.52 \\
          \textbf{w/o} SQL Selection Scaling & 77.95 & 66.59 & 62.76 & 73.08 & -1.82 \\
          \bottomrule
\end{tabular}
\vspace{-2mm}
\end{table*}

\subsection{Main Results (RQ1)}
As presented in Table \ref{table:main_bird}, our Agentar-Scale-SQL framework establishes a new SOTA on the BIRD benchmark, achieving 81.67\% execution accuracy (EX) and 77.00\% R-VES on the test set.
This result surpasses the prior SOTA (AskData + GPT-4o) by 0.79\% in test EX. 
It is worth noting that these approaches employ powerful models, such as Gemini 2.5 Pro and Claude 4.

Notably, the performance gain is even more pronounced when compared to single-trained models; Agentar-Scale-SQL outperforms the strongest single model (Gemini-SQL) by a substantial 5.54\% in test EX. These results empirically validate the effectiveness of our orchestrated scaling strategy.

\subsection{Ablation Study (RQ2)}
Next, we study the effectiveness of each module in Agentar-Scale-SQL by comparing Agentar-Scale-SQL with its variants without the key module. The results are listed in Table \ref{table:ablation}.

\noindent \textbf{Agentar-Scale-SQL w/o Task Understanding.}
In this variant, the overall EX improves by 0.45, showing that the task-understanding module resolves ambiguities—both in the values required for SQL clauses and in the phrasing of the questions themselves.

\noindent \textbf{Agentar-Scale-SQL w/o Reasoning SQL Generator.}
Removing the reasoning SQL generator leads to the most substantial performance drop, with total accuracy decreasing by 4.89 points. 
This result strongly demonstrates the effectiveness of the intrinsic scaling approach. 
It is essential for generating accurate SQL logic and aligning with the target data’s preferences, which is an indispensable asset in solving complex problems.

\noindent \textbf{Agentar-Scale-SQL w/o ICL SQL Generator.}
When the ICL SQL generator is excluded, the total accuracy falls by 3.78 points, the second-largest drop observed. Notably, the performance on challenging questions plummets from 64.14 to 55.86. This highlights the complementary nature of our two generators. The ICL generator excels at leveraging contextual examples to construct complex queries, providing an effective alternative pathway to a correct solution. The parallel scaling using two complementary generators ensures a diverse and high-quality pool of candidate SQL queries, which is crucial for achieving high performance.

\noindent \textbf{Agentar-Scale-SQL w/o Iterative Refinement.}
We also analyzed the contribution of the iterative refinement module via an ablation study. Its removal caused a 0.52-point drop in performance, which we attribute to the module's ability to polish the SQL and correct syntactic and semantic errors.

\noindent \textbf{Agentar-Scale-SQL w/o SQL Selection Scaling.}
Agentar-Scale-SQL w/o SQL selection scaling denotes that we employ self-consistency (i.e., majority voting based on execution results) to select the best SQL.
We can observe that our method outperforms the self-consistency baseline by 1.82 points in EX. This is because the most frequently executed result is not necessarily correct.
Our selection strategy, which likely incorporates more signals than simple frequency, proves to be a more effective and robust method for identifying the correct query.

\subsection{Analysis of Generator Components (RQ3)}
\begin{figure}[htbp]
    \centering
    \includegraphics[width=0.99\columnwidth]{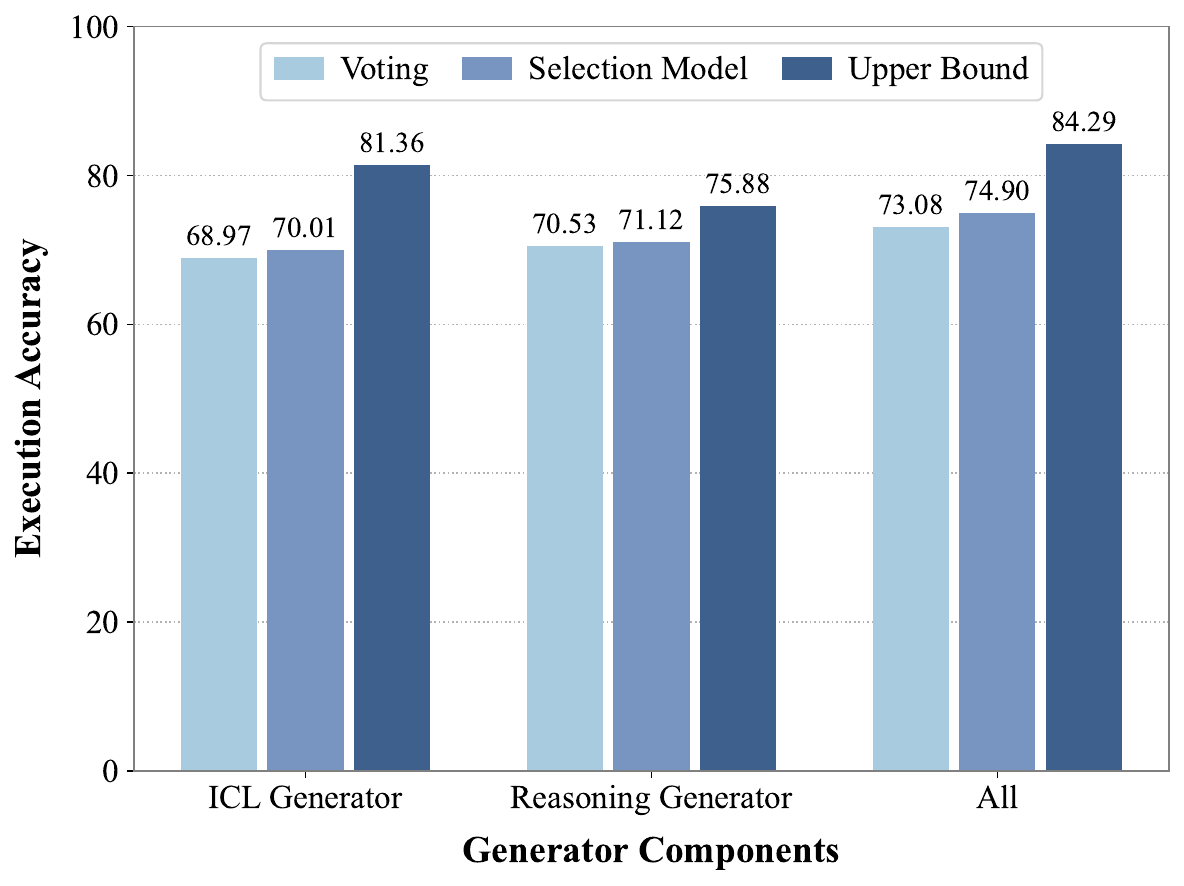}
    \caption{Execution accuracy of voting, selection model, and upper bound across generator components.}
    \label{fig:ex_components}

\end{figure}

\begin{figure}[htbp]
    \centering
    \includegraphics[width=0.99\columnwidth]{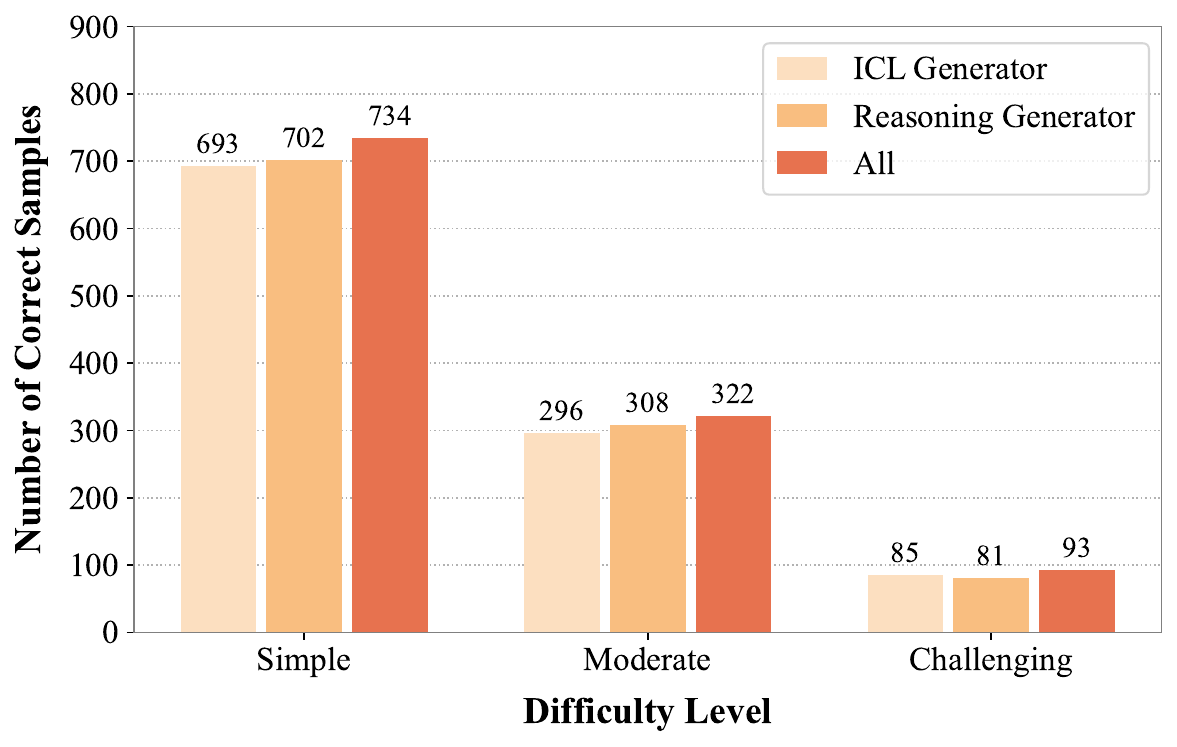}
    \caption{Number of correct samples by difficulty level for ICL, reasoning, and combined generators.}
    \label{fig:correct_cnt}

\end{figure}

As shown in Figure \ref{fig:ex_components}, the ICL generator achieves a notably higher upper bound accuracy (81.36\%) than the reasoning generator (75.88\%), indicating its strong potential for generating correct queries. However, combining their outputs (All) achieves the highest overall upper bound of 84.29\%. This synergistic gain is explained by their complementary nature, as illustrated in Figure \ref{fig:venn}. While they share a large set of correct solutions, they also uniquely solve 47 and 12 samples, respectively. This expanded coverage holds across all difficulty levels (Figure \ref{fig:correct_cnt}).
A breakdown by difficulty reveals that the reasoning generator holds an edge on simple and moderate tasks, while the ICL generator proves more effective in challenging problems.

Ultimately, this richer and more diverse candidate pool allows our final selection strategy to achieve its peak accuracy of 74.90\%, demonstrating the crucial role of the dual-generator approach.

\begin{figure}[htbp]
\centering
\begin{tikzpicture}[scale=0.6] 
    \def\radius{2.25cm}
    \coordinate (Apos) at (0,0);
    \coordinate (Bpos) at (2.5,0);

    \fill[fill={rgb,255:red,240; green,194; blue,132}] (Apos) circle (\radius);
    \fill[fill={rgb,255:red,168; green,203; blue,223}] (Bpos) circle (\radius);

    \begin{scope}
        \clip (Apos) circle (\radius);
        \fill[fill={rgb,255:red,204; green,199; blue,178}] (Bpos) circle (\radius);
    \end{scope}

    \node[anchor=east, align=center, font=\small] at (-2.1, 0) {ICL \\ Generator};
    \node[anchor=west, align=center, font=\small] at (4.6, 0) { Reasoning \\ Generator};
    
    \node[font=\small, text=black] at (-1, 0) {47};
    \node[font=\small, text=black] at (3.5, 0) {12};
    \node[font=\small, text=black] at (1.25, 0) {1090};

\end{tikzpicture}
\caption{Shared and unique correct samples between ICL and reasoning generators.}
\label{fig:venn}

\end{figure}

\subsection{Impact of the Candidate Number (RQ4)}
\begin{figure}[htbp]
    \centering
    \includegraphics[width=0.98\columnwidth]{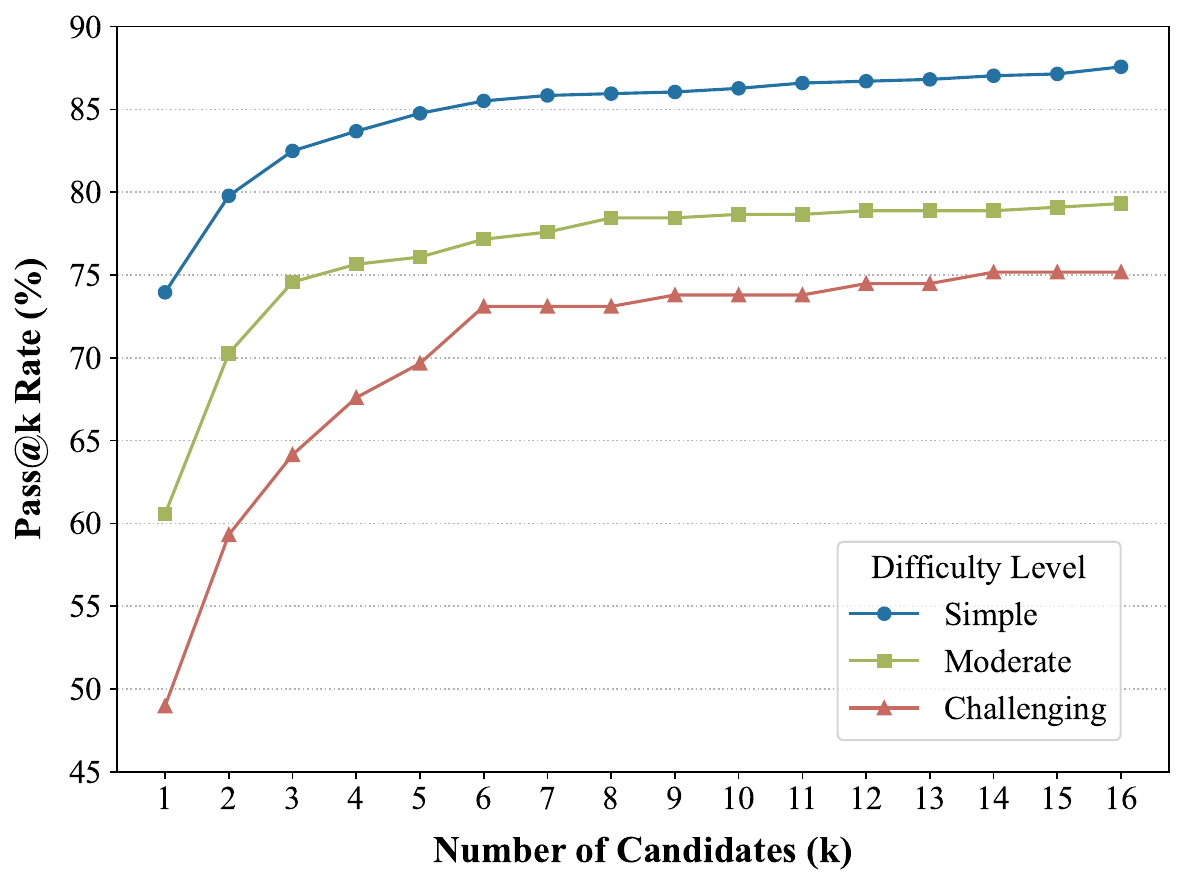}
    \caption{Pass@k rate with varying candidate numbers.}
    \label{fig:pass@k}
\vspace{-2mm}
\end{figure}

Finally, we investigate the impact of the candidate number by varying it from 1 to 16.
The results are depicted in Figure \ref{fig:pass@k}, showing that increasing the number of candidates consistently boosts the Pass@k rate across all difficulty levels. The improvement is most significant for challenging queries and is most substantial when increasing the candidates up to 8, after which the gains diminish. This validates the effectiveness of our parallel scaling strategy.

\section{Conclusion}
We introduced Agentar-Scale-SQL, a novel framework that significantly improves Text-to-SQL performance by synergistically combining internal, sequential, and parallel scaling strategies. Our method achieves SOTA results on the challenging BIRD benchmark, demonstrating an effective path toward human-level accuracy.
We release codes and models to support future research in this area.

Building on the success of enhancing intelligence through Test-Time Scaling, we are pioneering our next endeavor: Exercise-Time Scaling. We will empower a new generation of agents to learn through action and evolve from experience.

\clearpage

\section*{Limitations}
Despite its effectiveness, Agentar-Scale-SQL's reliance on orchestrated test-time scaling introduces several key limitations. The framework's primary drawback is its substantial computational overhead and high latency due to the multiple LLM calls for generation, refinement, and selection, making it less suitable for real-time applications. 
In our commercial practice (developing B2B ChatBI products), we observe that enterprise clients prioritize Accuracy above all else. In decision-making scenarios, a hallucinatory SQL query is unacceptable, whereas a latency of several seconds is often tolerable for complex analytics generation. Agentar-Scale-SQL is designed specifically to bridge the "last mile" of accuracy for these production requirements.
Furthermore, its performance is fundamentally bounded by the capabilities of the underlying base LLM, and it is susceptible to cascading errors where a failure in an early stage, such as task understanding, can compromise the entire process. 

\bibliography{custom}

\clearpage

\appendix

\section{BIRD Leaderboard: Bar Chart and Snapshot}
\begin{figure}[htbp] 
    \centering

    \begin{subfigure}[b]{\columnwidth}
        \centering
        \includegraphics[width=\textwidth]{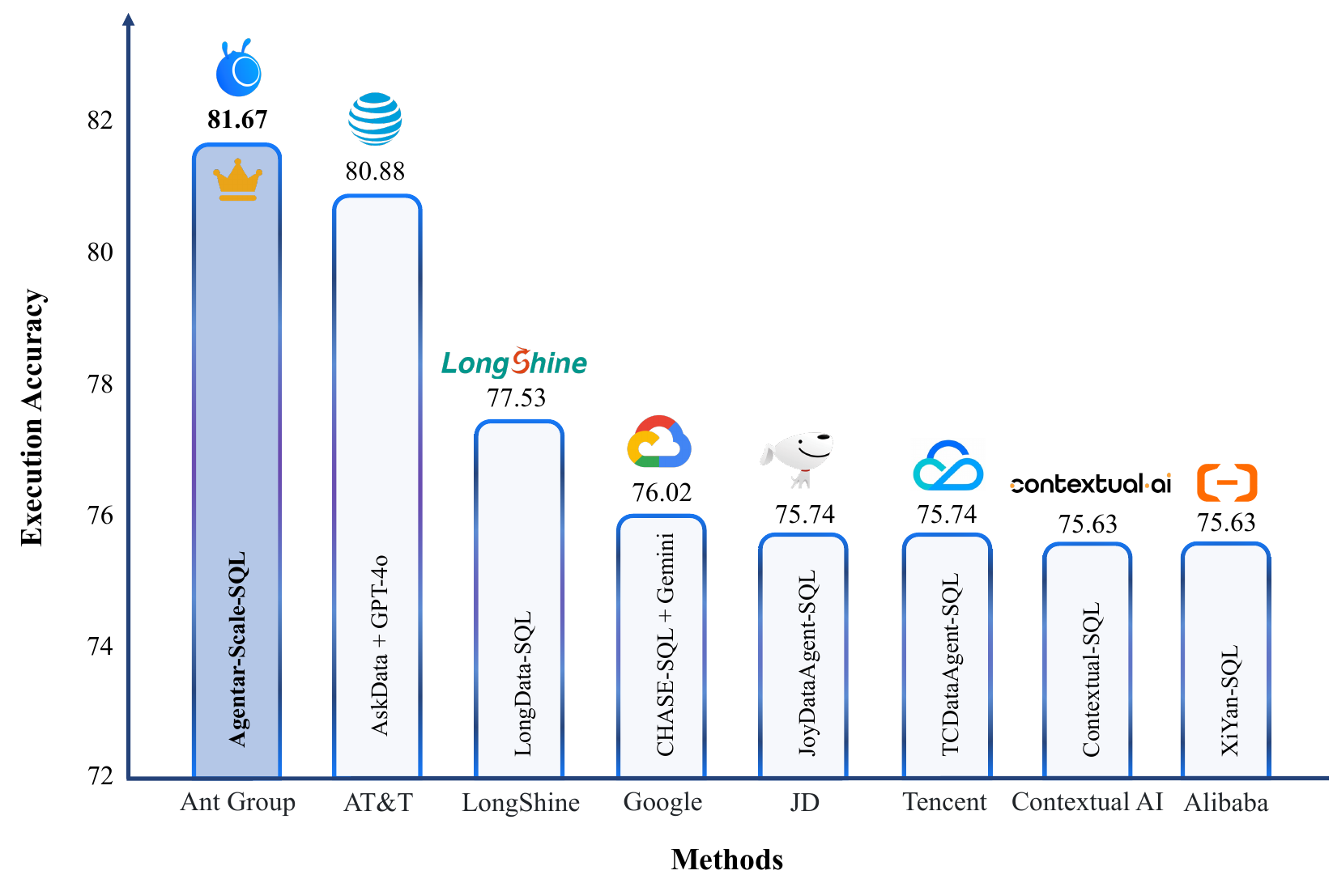}
        \caption{EX}
        \label{fig:ex}
    \end{subfigure}
    \hfill

    \begin{subfigure}[b]{\columnwidth}
        \centering
        \includegraphics[width=\textwidth]{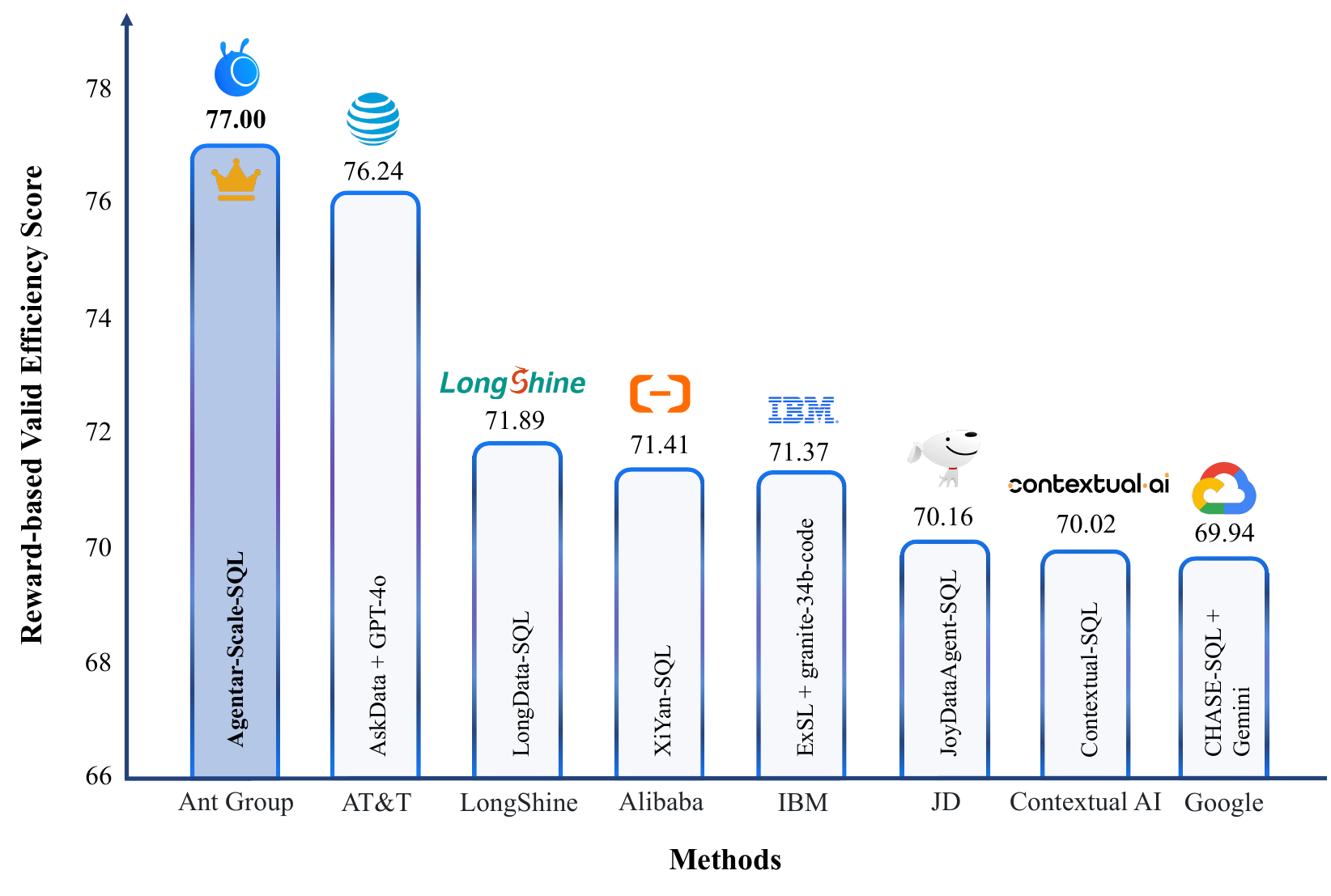}
        \caption{R-VES}
        \label{fig:r-ves}
    \end{subfigure}

    \caption{BIRD leaderboard (as of September 28, 2025): EX and R-VES performance comparison.}
    \label{fig:leaderboard_bar_chart}
\end{figure}

\begin{figure*}[htbp] 
    \centering
    \includegraphics[width=\textwidth]{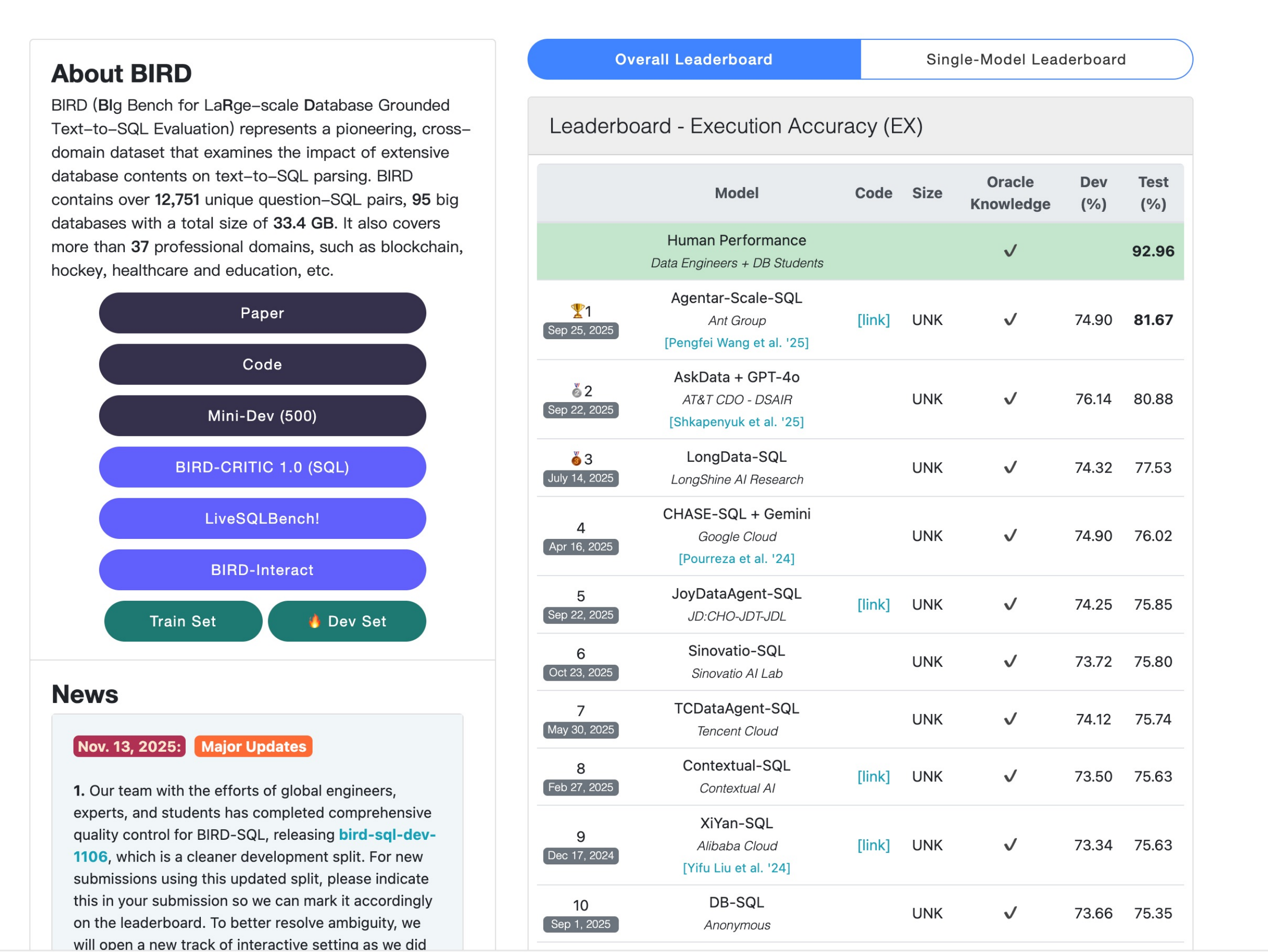}
    \caption{A snapshot of the BIRD benchmark's dynamic leaderboard as of November 27, 2025.}
    \label{fig:leaderboard}
\end{figure*}

We present the BIRD rankings as a bar chart (Figure \ref{fig:leaderboard_bar_chart}) and a snapshot (Figure \ref{fig:leaderboard}).

\section{Additional Results on the BIRD Test Set}

\begin{table}[h!]
\centering
\caption{Additional results on the BIRD test set.}
\label{tab:eval_results}
\resizebox{\linewidth}{!}{
\begin{tabular}{lcccc}
\toprule
\textbf{Metric} & \textbf{Simple} & \textbf{Moderate} & \textbf{Challenging} & \textbf{Total} \\
\midrule
Count & 949 & 555 & 285 & 1789 \\
\midrule
EX & 86.83 & 78.20 & 71.23 & 81.67 \\
Soft F1 & 87.28 & 79.41 & 73.58 & 82.65 \\
R-VES & 81.90 & 73.23 & 68.03 & 77.00 \\
\bottomrule
\end{tabular}}
\end{table}

\section{Definition of Evaluation Metrics}
This section provides detailed definitions for the three primary metrics \citep{DBLP:conf/nips/LiHQYLLWQGHZ0LC23} used to evaluate the performance of our text-to-SQL framework: Execution Accuracy (EX), Reward-based Valid Efficiency Score (R-VES), and the Soft F1-Score.

\subsection{Execution Accuracy (EX)}
Execution Accuracy (EX) is a strict binary metric that assesses whether the SQL query generated by the model produces the exact same result set as the ground-truth SQL query. 

For each question, the predicted SQL is executed against the database. The resulting table is then compared to the table generated by executing the official ground-truth SQL. A prediction is considered correct (score of 1) only if the two result sets are identical. Any deviation, including differences in row or column order, mismatched values, or SQL execution errors, results in a score of 0.

The final EX score for the entire evaluation set is the average of these binary scores, calculated as:
\begin{equation}
    \text{EX} = \frac{1}{N} \sum_{i=1}^{N} \text{score}_i
\end{equation}
where $N$ is the total number of questions in the evaluation set, and $\text{score}_i$ is 1 if the predicted query for question $i$ is correct, and 0 otherwise.

\subsection{Reward-based Valid Efficiency Score (R-VES)}
The Reward-based Valid Efficiency Score (R-VES) is designed to measure the execution efficiency of a correctly generated SQL query relative to its ground-truth counterpart. This metric is calculated only for queries that pass the Execution Accuracy (EX) evaluation.

The score is based on a reward function that considers the ratio of execution times between the predicted query ($T_{\text{pred}}$) and the ground-truth query ($T_{\text{gold}}$). The reward for a single valid query is defined as:
\begin{equation}
    \text{Reward}_i = 
    \begin{cases} 
        1 & \text{if } T_{\text{pred}_i} < T_{\text{gold}_i} \\
        \frac{2}{1 + (T_{\text{pred}_i} / T_{\text{gold}_i})} & \text{if } T_{\text{pred}_i} \geq T_{\text{gold}_i}
    \end{cases}
\end{equation}
This function assigns a full reward of 1 if the predicted query is more efficient than the ground truth. If it is slower, the reward decreases as the execution time ratio increases, penalizing inefficient queries.

The final R-VES is the average reward over all validly executed queries:
\begin{equation}
    \text{R-VES} = \frac{1}{N_{\text{valid}}} \sum_{i=1}^{N_{\text{valid}}} \text{Reward}_i
\end{equation}
where $N_{\text{valid}}$ is the total number of queries that passed the EX evaluation. To ensure stability, execution times are measured with a timeout, and the evaluation is repeated multiple times, with the highest score being reported.

\subsection{Soft F1-Score}
The Soft F1-Score offers a more lenient evaluation than EX by comparing the content similarity of the result tables produced by the predicted and ground-truth queries. This metric is insensitive to column order and robust to missing values.

The calculation proceeds by comparing the tables on a row-by-row, cell-by-cell basis. For each row in the predicted table and the ground-truth table, we compute the following three quantities:
\begin{itemize}
    \item \textbf{Matched ($tp_{\text{row}}$)}: The number of common cell values between a predicted row and its best-matching ground-truth row.
    \item \textbf{Pred\_only ($fp_{\text{row}}$)}: The number of cell values present in the predicted row but not in its matched ground-truth row.
    \item \textbf{Gold\_only ($fn_{\text{row}}$)}: The number of cell values present in the ground-truth row but not in its matched predicted row.
\end{itemize}
These row-level counts are then aggregated across all rows to compute the total True Positives (tp), False Positives (fp), and False Negatives (fn) for the entire table:
\begin{align*}
    \text{tp} &= \sum_{\text{all rows}} tp_{\text{row}} \\
    \text{fp} &= \sum_{\text{all rows}} fp_{\text{row}} \\
    \text{fn} &= \sum_{\text{all rows}} fn_{\text{row}}
\end{align*}
Finally, standard Precision, Recall, and F1-Score are calculated using these aggregated values:
\begin{align}
    \text{Precision} &= \frac{\text{tp}}{\text{tp} + \text{fp}} \\
    \text{Recall} &= \frac{\text{tp}}{\text{tp} + \text{fn}} \\
    \text{Soft F1-Score} &= 2 \cdot \frac{\text{Precision} \cdot \text{Recall}}{\text{Precision} + \text{Recall}}
\end{align}

\section{Discussion of Schema Linking}
Given the universal and pluggable architecture of our Agentar-Scale-SQL framework, we did not develop a built-in schema linking strategy for the BIRD benchmark. The framework is explicitly designed to be modular, facilitating the seamless integration of components like a schema linker. This approach is particularly advantageous for large-scale databases with numerous tables, as a dedicated schema linking module can be easily incorporated as needed, without altering the core system.

\section{Extended Analysis: Scaling, Efficiency, and Comparisons}
\label{app:extended_analysis}

In this section, we provide further elaboration on the core contributions of our method regarding test-time scaling orchestration, a detailed comparison of scaling capabilities against SOTA baselines, and a quantitative analysis of computational costs versus utility.

\subsection{Orchestrated Test-Time Scaling}
Our work distinguishes itself from naive combinations of existing modules through a systematic orchestration of Test-Time Scaling across three distinct dimensions.
Unlike simple stacking, we introduce specific designs to maximize synergy, such as \textit{Diverse Synthesis} (combining breadth via ICL with depth via RL-tuned reasoning) and \textit{Tournament Selection} (an RL-enhanced pairwise judge). This orchestration allows Agentar-Scale-SQL to validate the ``Scaling Hypothesis'' in Text-to-SQL, outperforming the strongest single model (Gemini-SQL) by 5.54\% and the previous SOTA (AskData + GPT-4o) by 0.79\%.

\subsection{Comparison of Scaling Capabilities}
Table~\ref{tab:scaling_comparison} categorizes existing SOTA methods based on their supported scaling dimensions. While most frameworks support only one or two dimensions, our method is the first to fully integrate all three, offering a comprehensive optimization space.

\begin{table*}[h]
    \centering
    \caption{Comparison of scaling capabilities across SOTA methods. Our method uniquely supports Internal, Sequential, and Parallel scaling simultaneously.}
    \label{tab:scaling_comparison}
    \renewcommand{\arraystretch}{1.2}
    \begin{tabular}{l c c c}
        \toprule
        \textbf{Methods} & \textbf{\makecell{Internal \\ Scaling}} & \textbf{\makecell{Sequential \\ Scaling}} & \textbf{\makecell{Parallel \\ Scaling}} \\
        \midrule
        \multicolumn{4}{l}{\textit{Overall}} \\
        XiYan-SQL & - & \checkmark & \checkmark \\
        Contextual-SQL & - & - & \checkmark \\
        CHASE-SQL + Gemini & - & \checkmark & \checkmark \\
        AskData + GPT-4o & - & - & \checkmark \\
        \midrule
        \multicolumn{4}{l}{\textit{Single Trained Models}} \\
        Arctic-Text2SQL-R1-32B & \checkmark & - & - \\
        Databricks RLVR 32B & \checkmark & - & - \\
        Gemini-SQL & \checkmark & - & - \\
        \midrule
        \textbf{Agentar-Scale-SQL (Ours)} & \textbf{\checkmark} & \textbf{\checkmark} & \textbf{\checkmark} \\
        \bottomrule
    \end{tabular}
\end{table*}

\subsection{Computational Cost vs. Practical Utility}
\begin{table*}[h] 
    \centering
    \caption{Comparison of inference costs (estimated number of LLM calls).}
    \label{tab:cost_comparison}
    \resizebox{\textwidth}{!}{
    \begin{tabular}{l c c c c c}
        \toprule
        \textbf{Methods} & \textbf{\makecell{Task \\ Understanding}} & \textbf{Generation} & \textbf{Refinement} & \textbf{Selection} & \textbf{Total Calls} \\ 
        \midrule
        \multicolumn{6}{l}{\textit{Overall}} \\
        Contextual-SQL & - & 1024 & - & 1024 & 2048 \\
        CHASE-SQL + Gemini & 32 & 16 & 32 & 10 & 297 \\
        \midrule
        \multicolumn{6}{l}{\textit{Single Trained Models}} \\
        Sophon-Text2SQL-32B & - & 8 - 32 & - & - & 8 - 32 \\
        Databricks RLVR 32B & - & 1 - 7 & - & - & 1 - 7 \\
        Gemini-SQL & - & 1 - 7 & - & - & 1 - 7 \\
        \midrule
        \textbf{Agentar-Scale-SQL (Ours)} & 1 & 17 & 1 - 51 & 1 - 561 & 20 - 630 \\
        \bottomrule
    \end{tabular}
    }
\end{table*}

While test-time scaling inherently introduces higher latency compared to single-pass models, this is a deliberate trade-off designed for high-stakes enterprise scenarios. In B2B ChatBI products, accuracy is paramount; a hallucinated query in a financial report is unacceptable, whereas a latency of 10--30 seconds is often tolerable for complex analytics. 

Furthermore, our framework offers high configurability. Users can dynamically adjust the number of candidates ($N$) or skip the refinement stage to reduce costs for simpler queries, bridging the ``last mile'' of accuracy only when necessary.

Table~\ref{tab:cost_comparison} provides an approximate comparison of inference costs using the number of LLM calls as a metric. Although our method involves more steps than single-pass models, it significantly optimizes the total calls compared to other multi-stage frameworks (e.g., Contextual-SQL) while achieving SOTA performance.

\section{Prompts}
We have listed a selection of prompts. For the complete list, please refer to the code repository.

\begin{itemize}
    \item The prompt for the Task Understanding.
    \item The prompt for the Reasoning Generator.
    \item The prompt for the ICL Generator with CoT prompting.
    \item The prompt for the SQL fixer.
    \item The prompt for the Reasoning Selector.
\end{itemize}

\begin{figure*}[t]
\centering
\begin{lstlisting}[style=prompt]
# Task Description
You are a text-to-SQL expert who is excellent in analysing text-to-SQL question. You are given ['Database Schema', 'Question', 'Evidence'], you need to **comprehensively analyse the question** by:
1. Identifying the database literals appeared in the question or the evidence. Database literal refers to a specific value that belongs to a column (e.g., "Japan", "Chinese Grand Prix"). These values are used in the WHERE clause to conduct effective filtering.
2. Generating a question skeleton. The question skeleton contains the question structure while ignoring the detailed database information (entity names, column names).

# Guideline for identifying database literals
1. Taking both the question and the evidence into account.

# Guideline for generating a question skeleton
1. Replace **all the database literals** in the question with the database column it belongs to.
2. Replace **all the database columns** with the placeholder <COLUMN>.
3. Keep SQL keywords such as "average", "total", "difference", "count" in the question skeleton.

# Example of the question mapping to skeleton
Question: Name movie titles released in year 1945. Sort the listing by the descending order of movie popularity.
Skeleton: Name <COLUMN> released in <YEAR>. Sort the listing by the descending order of <COLUMN>.

Question: In August of 1996, how many orders were placed by the customer with the highest amount of orders?
Skeleton: In <MONTH> of <YEAR>, how many <COLUMN> were placed by the <COLUMN> with the the highest amount of <COLUMN>?

Question: Calculate the total production for each product which were supplied from Japan.
Skeleton: Calculate the total <COLUMN> for each <COLUMN> which were supplied from <COUNTRY>.

Question: Calculate the difference in the average number of low-priority orders shipped by truck in each month of 1995 and 1996.
Skeleton: Calculate the difference in the average number of <COLUMN> in each month of <YEAR> and <YEAR>.

# About Output
**Do not return explanations or reasoning process.**
The output should be formatted as a JSON instance that conforms to the JSON schema below.

As an example, for the schema {"properties": {"foo": {"title": "Foo", "description": "a list of strings", "type": "array", "items": {"type": "string"}}}, "required": ["foo"]}
the object {"foo": ["bar", "baz"]} is a well-formatted instance of the schema. The object {"properties": {"foo": ["bar", "baz"]}} is not well-formatted.

Here is the output schema:
```
{"properties": {"database_literals": {"description": "The database literals extracted from the question and evidence.", "items": {"type": "string"}, "title": "Database Literals", "type": "array"}, "question_skeleton": {"description": "the generated question skeleton", "title": "Question Skeleton", "type": "string"}}, "required": ["database_literals", "question_skeleton"]}
```
\end{lstlisting}
\caption{The system prompt for the Task Understanding.}
\label{fig:prompt_system_task_understanding}
\end{figure*}

\begin{figure*}[t]
\centering
\begin{lstlisting}[style=prompt]
# Database Schema
{Database Schema}

# Question
{Question}

# Evidence
{Evidence}

Begin! Take a deep breath and think logically.
[no prose][output json Only]
\end{lstlisting}
\caption{The user prompt for the Task Understanding.}
\label{fig:prompt_user_task_understanding}
\end{figure*}

\begin{figure*}[t]
\centering
\begin{lstlisting}[style=prompt]
Task Overview:
You are a data science expert. Below, you are provided with a database schema and a natural language question. Your task is to understand the schema and generate a valid SQL query to answer the question.

Database Engine:
{dialect}

Database Schema:
{Database Schema}
This schema describes the database's structure, including tables, columns, primary keys, foreign keys, and any relevant relationships or constraints.
Matched contents:
{matched_contents}
Matched contents presents values related to the question, together with their source table and column, for your reference in SQL generation.
Question:
{evidence}
{question}

Instructions:
- If Matched contents is provided, you can use it as reference when generating the SQL query.
- Make sure you only output the information that is asked in the question. If the question asks for a specific column, make sure to only include that column in the SELECT clause, nothing more.
- The generated query should return all of the information asked in the question without any missing or extra information.
- Before generating the final SQL query, please think through the steps of how to write the query.

Output Format:
In your answer, please enclose the generated SQL query in a code block:
```sql
-- Your SQL query
```

Take a deep breath and think step by step to find the correct SQL query.
\end{lstlisting}
\caption{The prompt for the Reasoning Generator.}
\label{fig:prompt_reasoning_generator}
\end{figure*}

\begin{figure*}[t]
\centering
\begin{lstlisting}[style=prompt]
1. **SELECT Clause:** 
    - Only select columns mentioned in the user's question. 
    - Avoid unnecessary columns or values.
2. **Aggregation (MAX/MIN):**
    - Always perform JOINs before using MAX() or MIN().
3. **ORDER BY with Distinct Values:**
    - Use `GROUP BY <column>` before `ORDER BY <column> ASC|DESC` to ensure distinct values.
4. **Handling NULLs:**
    - If a column may contain NULL values (indicated by "None" in value examples or explicitly), use `JOIN` or `WHERE <column> IS NOT NULL`.
    - When a field is sorted in ascending order, also apply a NOT NULL filter to it.
    - When using the MIN() function on a column, also include a WHERE clause to filter NULL values from that column.
5. **FROM/JOIN Clauses:**
    - Only include tables essential to answer the question.
6. **Strictly Follow evidences:**
    - Adhere to all provided evidences.
7. **Thorough Question Analysis:**
    - Address all conditions mentioned in the question.
8. **DISTINCT Keyword:**
    - Use `SELECT DISTINCT` when the question requires unique values (e.g., IDs, URLs). 
    - Refer to column statistics ("Value Statics") to determine if `DISTINCT` is necessary.
9. **Column Selection:**
    - Carefully analyze column descriptions and evidences to choose the correct column when similar columns exist across tables.
10. **String Concatenation:**
    - Never use `|| ' ' ||` or any other method to concatenate strings in the `SELECT` clause. 
11. **JOIN Preference:**
    - Prioritize `INNER JOIN` over nested `SELECT` statements.
12. **SQLite Functions Only:**
    - Use only functions available in SQLite.
13. **Date Processing:**
    - Utilize `STRFTIME()` for date manipulation (e.g., `STRFTIME('%Y', SOMETIME)` to extract the year).
14. **Formatting:**
    - Pay close attention to any formatting requirements in the question, such as specific decimal places or percentage representation. These are not just suggestions; they are critical parts of the final answer and must be implemented using appropriate SQL functions (e.g., ROUND() and multiplying by 100).
    - Use `ROUND()` to round the result to a specific number of decimal places.
    - Use `* 100` to convert a fraction to a percent (%).
\end{lstlisting}
\caption{The database admin instruction.}
\label{fig:prompt_database}
\end{figure*}

\begin{figure*}[t]
\centering
\begin{lstlisting}[style=prompt]
# Your Role
You are an experienced database expert. As a SQL expert, formulate a precise SQL query to resolve the user's question. You must carefully analyze the database schema, the question's intent, and any provided examples or experience to ensure correctness and efficiency.
The database stores millions of cell values, you need to write a SQL query that can be executed in **less than 20.0 seconds**. Otherwise the business will be effected and lose millions of dollars.

# Database Engine
{dialect}

# Database Admin Instruction
{Database Admin Instruction}

# Input Information
You are provided with ['Examples', 'Database Schema', 'Question', 'Evidence'].
**Examples** are the similar question-SQL pairs which are annotated by other experienced database experts.
**Database schema** are the database structure. It contains table names, column names, column types, primary key and foreign key relationship. It also contains some relevant database cell values to the question.
**Question** is your task to answer.
**Evidence** contains expert experience about the question. It is **very important reference** to answer the question.

# About Output
Please return your analysis and then write the SQL query in a code block:
1. Decompose the question and evidence, then map the question and evidence to the schema.
2. [Important] If the SQL query does not require any subqueries, write the complete SQL query directly.
3. [Important] If the SQL query does require subqueries, **first write the subquery (or subqueries), then write the main (outer) query that uses them**.
4. You need to enclose the final SQL query in a code block using the following format:
```sql

```

## Here is an example about the process of analysing question and generating SQL query
### Question
How many users have rated the most popular movie?
### Evidence

### Answer
Step 1: Decompose the question.
- We are asked to count the number of users who have rated the most popular movie.
- This requires us to first identify the most popular movie (a subquery is needed), then count the ratings for that movie.

Step 2: Write the subquery to find the most popular movie.
```sql
SELECT movie_id
FROM movies
ORDER BY movie_popularity DESC
LIMIT 1
```

Step 3: Write the main query to count the ratings for that movie.
```sql
SELECT COUNT(rating_id)
FROM ratings
WHERE movie_id = (
    SELECT movie_id
    FROM movies
    ORDER BY movie_popularity DESC
    LIMIT 1
)
```
\end{lstlisting}
\caption{The system prompt for the ICL Generator with CoT prompting.}
\label{fig:prompt_system_icl_generator_cot}
\end{figure*}

\begin{figure*}[t]
\centering
\begin{lstlisting}[style=prompt]
# Examples
{Examples}

# Database Schema
{Database Schema}

# Question
{Question}

# Evidence
{Evidence}

Begin! Take a deep breath and think logically.
\end{lstlisting}
\caption{The user prompt for the ICL Generator with CoT prompting.}
\label{fig:prompt_user_icl_generator_cot}
\end{figure*}

\begin{figure*}[t]
\centering
\begin{lstlisting}[style=prompt]
# About Role
You are a database expert excellent in writing SQL query. Your task is to correct a wrong SQL query.

# SQL Engine
{dialect}

# Input Information
- Database schema: Contains the full structure of the database.
- Question: The natural language question that needs to be answered.
- Evidence: Key information extracted from the question and/or database that helps answer it.
- Original SQL: The SQL query that was previously executed but resulted in an error or empty result.
- Execution Result: The outcome of executing the original SQL. It can be a database error or an empty result.

# Task Description:
Correct the SQL query accordingly:
   - Fix any syntax errors.
   - Adjust filtering conditions or column references based on evidence.
   - Remove unnecessary JOINs if they lead to empty intersections.
   - Ensure the corrected SQL still corresponds one-to-one with the targets and conditions in the question.

# About Output
Directly output the SQL query in the code block:
```sql
```
\end{lstlisting}
\caption{The system prompt for the SQL Fixer.}
\label{fig:prompt_system_sql_fixer}
\end{figure*}

\begin{figure*}[t]
\centering
\begin{lstlisting}[style=prompt]
# Database Schema
{Database Schema}

# Question
{Question}

# Evidence
{Evidence}

# Original SQL
{Original SQL}

# Execution Result
{Execution Result}

Begin! Take a deep breath and think logically.
[no prose][output result Only]
\end{lstlisting}
\caption{The user prompt for the SQL Fixer.}
\label{fig:prompt_user_sql_fixer}
\end{figure*}

\begin{figure*}[t]
\centering
\begin{lstlisting}[style=prompt]
You are an advanced SQL evaluation assistant. Your task is to evaluate multiple SQL query candidates in response to a schema-related question. For each candidate, you will be provided with the SQL query and its execution result. Carefully analyze the query and its result for correctness, completeness, and relevance to the question and schema. Select the candidate that best answers the question, and briefly explain your reasoning.

# SQL Engine
sqlite

# About Input
The user will provide you with ['Database Schema', 'Matched contents', 'Evidence', 'Question', 'SQL Candidates']. Use this information to evaluate which SQL candidate best answers the question.

# About Output
- You are encouraged to provide your reasoning process before giving the final answer.
- For the final answer, output only the selected SQL label, wrapped clearly 
    within `\\boxed{}` for easy identification and extraction.
- Example: If you select SQL candidate 2, output: `\\boxed{2}`
\end{lstlisting}
\caption{The system prompt for the Reasoning Selector.}
\label{fig:prompt_system_reasoning_selector}
\end{figure*}

\begin{figure*}[th]
\centering
\begin{lstlisting}[style=prompt]
# Database Schema:
{Database Schema}
# Matched contents:
Matched contents present values related to the question, together with their source table and column, for your reference in SQL selection.
{matched_contents}
# Evidence:
{evidence}
# Question:
{question}
# SQL Candidates:
{candidates}
\end{lstlisting}
\caption{The user prompt for the Reasoning Selector.}
\label{fig:prompt_user_reasoning_selector}
\end{figure*}

\section{Further Work}
Agentar-Scale-SQL marks a significant milestone on our journey toward Artificial General Intelligence (AGI) and Artificial Superintelligence (ASI). By leveraging \emph{Orchestrated Test-Time Scaling}, we have substantially advanced the state-of-the-art in Text-to-SQL. Looking ahead, we plan to explore the following directions:

\begin{itemize}
    \item \textbf{Self-Exploration:} We will enable the agent to autonomously explore and accumulate experience in an offline phase, thereby shifting the computational burden from Test-Time to a pre-computation phase we term Exercise-Time.

    \item \textbf{Agentic SQL:} We aim to evolve our current workflow-based approach into a fully autonomous agent, moving beyond predefined structures.

    \item \textbf{Generalization:} We intend to extend the \emph{Orchestrated Test-Time Scaling} methodology to a broader range of code generation and reasoning tasks.
\end{itemize}

\end{document}